\documentclass[preprint,12pt]{elsarticle}

\usepackage{lineno,hyperref}
\modulolinenumbers[5]
\usepackage{amssymb}
\usepackage{multirow}
\usepackage{rotating}
\usepackage{xcolor}
\usepackage{amsmath}
\usepackage{amsfonts}
\journal{AI in Medicine}

\begin{document}

\begin{frontmatter}



\title{A multi-stage machine learning model for diagnosis of esophageal manometry}


\author[label1]{Wenjun Kou\corref{corwk}}
\cortext[corwk]{Corresponding author}
\ead{w-kou@northwestern.edu}
\author[label1]{Dustin A. Carlson}
\author[label1]{Alexandra J. Baumann}
\author[label1]{Erica N. Donnan}
\author[label2]{Jacob M. Schauer}
\author[label3,label4]{Mozziyar Etemadi}
\author[label1]{John E. Pandolfino}
\address[label1]{Department of Medicine, Feinberg School of Medicine, Northwestern University, 676 North Saint Clair Street, 14th Floor, Chicago, IL 60611, USA}
\address[label2]{Department of Preventive Medicine, Feinberg School of Medicine, Northwestern University, 750 North Lake Shore Drive, 11th Floor, Chicago, IL 60611, USA}
\address[label3]{Department of Anesthesiology, Feinberg School of Medicine, Northwestern University, Chicago, IL 60611, USA}
\address[label4]{Department of Biomedical Engineering, McCormick School of Engineering, Northwestern University, Evanston IL 60201, USA}

\begin{abstract}
High-resolution manometry (HRM) is the primary procedure used to diagnose esophageal motility disorders. Its manual interpretation and classification, including evaluation of swallow-level outcomes and then derivation of a study-level diagnosis based on Chicago Classification (CC), may be limited by inter-rater variability and inaccuracy of an individual interpreter. We hypothesized that an automatic diagnosis platform using machine learning and artificial intelligence approaches could be developed to accurately identify esophageal motility diagnoses. Further, a multi-stage modeling framework, akin to the step-wise approach of the CC, was utilized to leverage advantages of a combination of machine learning approaches including deep-learning models and feature-based models. Models were trained and tested using a dataset comprised of 1741 patients’ HRM studies with CC diagnoses assigned by expert physician raters. In the swallow-level stage, three models based on convolutional neural networks (CNNs) were developed to predict swallow type and swallow pressurization (test accuracies of 0.88 and 0.93, respectively), and integrated relaxation pressure (IRP)(regression model with test error of 4.49 mmHg). At the study-level stage, model selection from families of the expert-knowledge-based rule models, xgboost models and artificial neural network(ANN) models were conducted. A simple model-agnostic strategy of model balancing motivated by Bayesian principles was utilized, which gave rise to model averaging weighted by precision scores. The averaged (blended) models and individual models were compared and evaluated, of which the best performance on test dataset is 0.81 in top-1 prediction, 0.92 in top-2 predictions. This is the first artificial-intelligence style model to automatically predict esophageal motility (CC) diagnoses from HRM studies using raw multi-swallow data and it achieved high accuracy. Thus, this proposed modeling framework could be broadly applied to assist with HRM interpretation in a clinical setting.
\end{abstract}

 \begin{graphicalabstract}
\includegraphics[scale = 0.5]{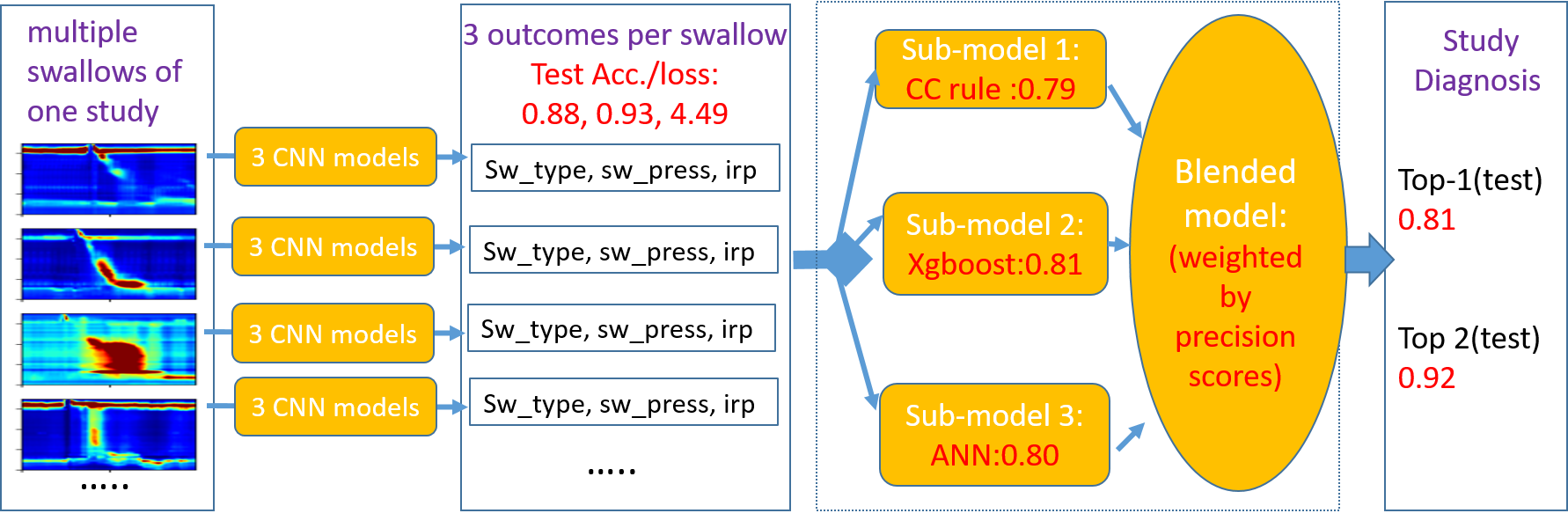}
 \end{graphicalabstract}
\begin{highlights}
\item A multi-stage artificial intelligence model for esophageal manometry was developed
\item Three convolutional neural network models on swallow type, pressurization and IRP were built
\item Study-level models to predict esophageal motility diagnoses were then developed

\item A model-agnostic approach of model balancing was proposed to develop blended models

\item The proposed framework is extensible for multi-modal tasks with multiple data sources
\end{highlights}


\end{frontmatter}


\section{Introduction}

High-resolution Manometry (HRM) is the primary method for clinical evaluation of esophageal motility disorders~\citep{kahrilas2015chicago,roman2016high,tolone2018high}. Its interpretation and classification is currently based on the Chicago Classification (CC), which categorizes each HRM study into an esophageal motility diagnosis based on a tree-like algorithm~\citep{kahrilas2015chicago}. The specific process of interpreting HRM relies on two stages using feature-based analysis algorithms. During the first stage, outcomes of each swallow are calculated via landmark identification and pre-defined metrics. Several swallow-level diagnoses are derived based on established rules, such as swallow type (normal, hypercontractile, weak, fragmented, failed, premature) and swallow pressurization (normal pressurization, pan-esophageal pressurization, and compartmentalized pressurization). Additionally, pressure at the esophagogastric junction (EGJ) after each swallow is measured using a metric, integrated relaxation pressure (IRP); the median IRP value of the test swallows is then applied to classify abnormal (elevated IRP) or normal EGJ outflow. During the second stage, the swallow-level diagnoses and outcomes are collected as inputs to derive the \textit{study} diagnosis (i.e. diagnosis of esophageal motility disorder) based on the established CC algorithm. A detailed list of label names in both the swallow-level categories and study-level categories can be found in Table.~\ref{table_data_1}. While the CC algorithm provides a framework for uniform esophageal motility interpretation, manual and subjective input are required for feature extraction. Hence, the interpretation of HRM is subject to inter-rater variability and, consequently, rater associated inaccuracy. A related study on outcome evaluation showed that inconsistent landmark identification is a major contributor to inter-rater disagreement~\citep{carlson2018inter}. Reliance on subjective experience also poses a challenge in medical training. A recent post-training evaluation survey from GI fellows showed that only 84 percent were confident in the interpretation of esophageal manometry~\citep{rao2015advanced}. Artificial intelligence (AI) or machine learning could potentially alleviate some of the issues associated with the clinical interpretation of HRM, evidenced by its success in other fields of medicine~\citep{miotto2018deep,hosny2018artificial,johnson2018artificial}. However, the current focus of machine learning in gastroenterology seemingly has been on the detection of intestinal malignancies or premalignant lesions from endoscopy videos or x-ray images~\citep{le2020application,ahmad2019artificial,ruffle2019artificial}. More recently, we developed an unsupervised machine learning model on esophageal swallow-level data for feature extraction to identify contraction patterns and separate swallow types. Wang et al. introduced a novel contractile vigor propagation graph attention network (CVP-GAT) model to recognize esophageal contraction patterns from high-resolution manometry images~\cite{wang2021attention}. While these studies demonstrated the capabilities of ML models to identify swallow types from HRM studies, the greater clinical value is derived from incorporating these swallow type classifications into a global esophageal motility diagnosis through integration with other features of the HRM study.

Machine learning, in a broad sense, can be classified into two types. One type is based on processed low-dimensional features/outcomes and focuses on the development and improvement of classification algorithms. The other type is directly based on large dimensional raw data, such as pressure measurements, to assign a classification label, often referred to as a deep learning model. Deep learning models often involve minimal data processing, but typically require a large number of observations to counter limitations inherent with machine learning with a large number of variables, often referred to as the curse of dimensionality~\citep{goodfellow2016deep}. Our previous efforts on data collection gave rise to a total dataset of thirty-two thousand distinct swallows, but less than three thousand studies~\citep{kou2021deep}. 
Hence, to develop an AI model on study-level diagnosis, we developed a multi-stage modeling framework that includes deep-learning models at the swallow-level stage and feature-based machine learning models at the study-level stage. In particular, at the study-level stage, a detailed model selection from various model families is conducted, which included models designed and augmented by expert knowledge. A simple strategy of model-agnostic model balancing motivated by Bayesian principles is proposed, which gives rise to model averaging weighted by precision scores. The averaged (or blended) models with various combinations of sub-models are compared and evaluated. We theorize that the proposed modeling framework could be easily extended to multi-modal tasks, such as diagnosis of esophageal patients based on clinical data from multiple procedures including esophageal manometry and functional luminal imaging probe panometry (FLIP)~\citep{carlson2019normal}.

\section{Materials and Methods}
\label{method_label}

\subsection{Overview of High-resolution Manometry data}
\begin{figure}[ht]
 \centering 
 \includegraphics[scale = 0.43]{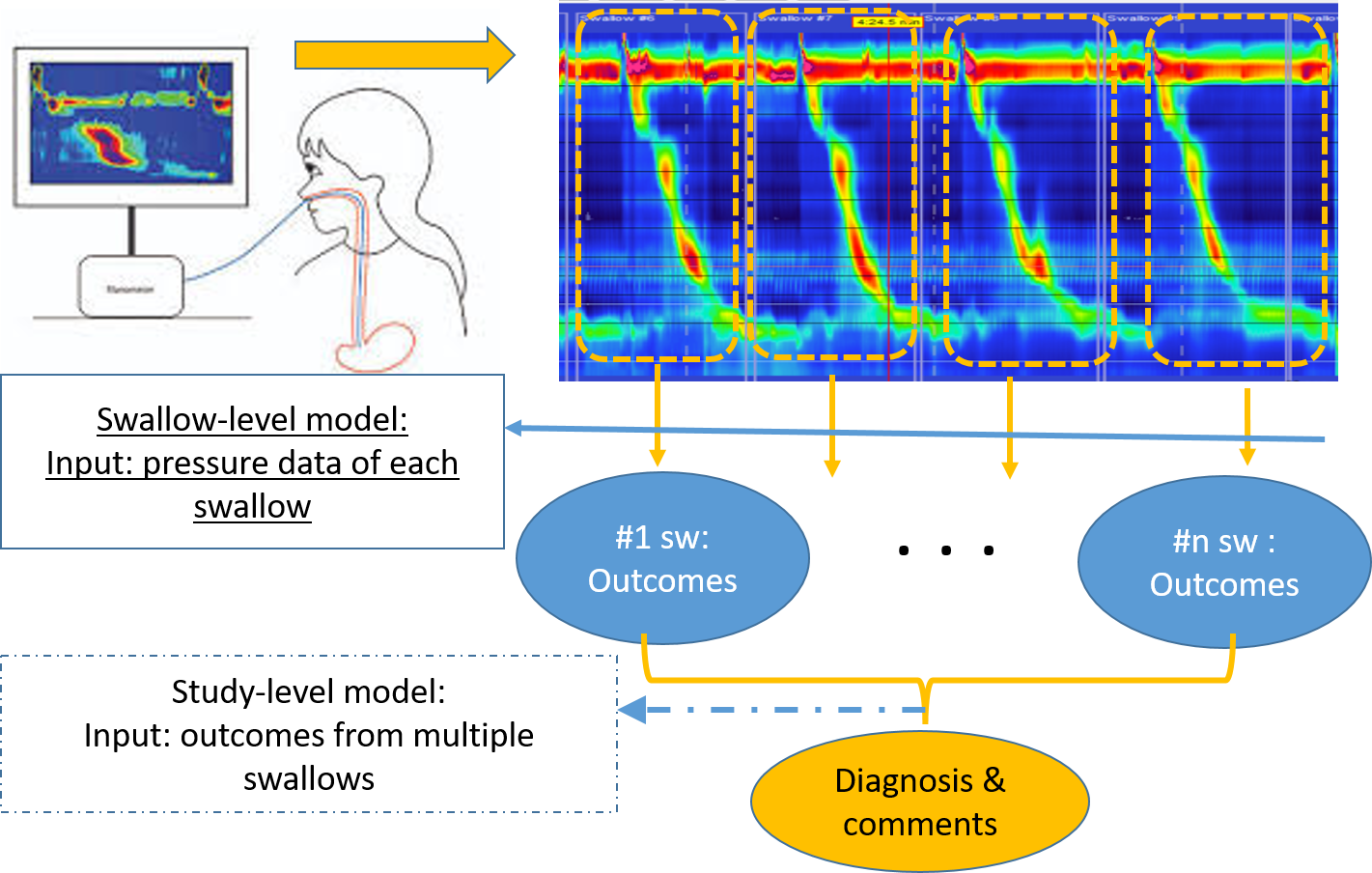}
 \caption{Illustration of high-resolution manometry procedure and associated information flow from raw data to final diagnosis in current clinical practices (reproduced with permission from~\citep{kou2021deep}).}
 \label{fig-overall}
\end{figure}

HRM is a routine procedure to evaluate esophageal functions. A typical HRM study includes multiple swallows of instructed food bolus, during which, associated pressure variations reflecting physiology of esophageal function are recorded by 36 pressure sensors that sample at frequency of 100Hz. The 36 pressure sensors (also referred to as channels) span from the upper esophageal sphincter to the stomach. The post-procedure evaluation starts with examining each swallow. Swallow-level data, comprising the pressure data within a 12 to 20 second window starting from swallow onset, is characterized independently to obtain various {\it swallow-level} features. Then based on then CC algorithm, those swallow-level features are then used to derive study-level labels as the diagnosis of the whole HRM study. An illustration of this multi-stage information flow is shown in Figure~\ref{fig-overall}.

The current manometry dataset were obtained at Northwestern Memorial Hospital with IRB approval. Specifically, data were collected on adult patients (age 18-89) presenting to the Esophageal Center of Northwestern for evaluation of esophageal symptoms between 2015 and 2019 who completed clinical HRM studies, and maintained in an esophageal motility registry were prospectively evaluated.  Patients with previous foregut surgery (including previous pneumatic dilation) or technically limited HRM studies were excluded. HRM data from 1,741 consecutively evaluated patients were included. There is some overlap with a previously described cohort~\citep{kou2021deep}. HRM data were processed and the dataset was restricted to contain HRM studies that satisfied the following criteria: 1) clearly labeled swallow type based on CC rules; 2) having 10 supine and 5 upright swallows with swallow type labels. The HRM studies were interpreted to assign labels of swallow type, pressurization, IRP, and esophageal motility diagnosis according to CC version 3.0, by an expert in HRM interpretation and one of the founders of the Chicago Classification and leads in subsequent updates (JEP). The labels were subsequently reviewed and confirmed by review of three additional experience raters (DAC, ED, AB) with any discrepancies adjudicated by consensus discussion of the group.

Based on distribution of study labels, the whole dataset was split to the three datasets (training, validation, and test) for both swallow-level and study-level models. Specifically, we started with the study-level split. We first computed the class distribution at the study level on the whole dataset (i.e. 1741 studies), and then conducted stratified random sampling to obtain study-level training/validation/test datasets with a 70/15/15 split. This procedure also generated training/validation/test datasets at the swallow level by assigning swallows to the same split as their corresponding study. For instance, swallows from all the studies in the study-level training dataset were assigned to the swallow-level training dataset. Table.~\ref{table_data_1} summarizes the distributions of study- and swallow-level labels across datasets. As it shows, the datasets on both the swallow level and the study level were constructed so that: 1) swallows within the same study were in the same dataset to avoid information leaking; and 2) sample distribution among training, validation, and test datasets was statistically similar.

\begin{table}[!htb]
 \caption{Sample distribution with respect to study-level categories (the Chicago Classification) and two swallow-level categories (swallow type and swallow pressurization). Note for category, \textit{Chicago Classification}, listed are the number of \textit{studies} in each dataset for the corresponding label. Also, Label \textit{IEM} includes the studies labeled as Fragmented Peristalsis (FRP) in original Chicago Classification, and Label \textit{T3A} includes those labeled as Distal Esophageal spasm (DES) originally~\citep{kahrilas2015chicago}.  For Categories, \textit{swallow type} and \textit{swallow pressurization}, listed are the number of \textit{swallows} in each dataset.}
 \centering
\begin{tabular}{c | c | p{3cm} | c |c | c }
 \hline \hline
 Category & label(id)  & label details &train $\#$ &valid. $\#$&test $\#$\\ [1ex]
 \hline
 Chicago  	  & ABC(0) &  Absent Contractility & 55 & 11 & 13 		\\
 classification	& T1A(1) &  Type1 Achalasia & 47 & 9 & 10 		\\
                & T2A(2) &   Type2 Achalasia & 93 & 18 & 18		\\
                & T3A(3) &   Type3 Achalasia & 64 & 13 & 15		\\
                & EGJOO(4) &   EGJ outflow obstruction & 207 & 40 & 48		\\
                & JES(5) &   Jackhammer esophagus &27 & 7 & 6		\\
                & NEM(6) &   Normal motility & 565 & 116 & 119 \\
                & IEM(7) &   Ineffective esophageal motility & 165 & 38 & 37	\\			      
 \hline
 swallow type	&N(0)& Normal &     10070 &     2083 &     2260 	\\
     		    &W(1)   & Weak &      1947 &      410 &      392	\\
			    &F(2)   & Failed &      4677 &     924 &     1022 		\\
			    &FR(3)   & Fragmented &       577 &      154 &      105	\\
			    &P(4)   & Premature &       507 &      99 &      112		\\
			    &H(5)   & Hypercontraction &       567 &      110 &      99		\\
 \hline
 swallow 		&NP(0)    & Normal pressurization &     15529 &    3201 &     3408		\\
 pressurization &CP(1)    & Compartmental pressurization  &     1641 &      328 &      277		\\
 	            &PEP(2)    & Panesophageal pressurization  &     1175 &      251 &      305	\\
 \hline
 \end{tabular}
 \label{table_data_1}
\end{table}

Similar to previous work on unsupervised modeling, down-sampling of raw pressure data was conducted to address dimensionality~\citep{kou2021deep}. In particular, the time duration for each swallow was set as 24 seconds and the data was down-sampled from 100Hz to 10Hz. This choice of time window and down-sampling procedure was chosen for several reasons. The 24-second time windows is sufficient to include pre-swallow and post-swallow data to retain physiologically-relevant information. The down-sampling approach, as evidenced in previous work, did not demonstrably sacrifice the $information$ associated with patterns of the esophageal swallow. However, unlike previous work~\citep{kou2021deep}, data augmentation was not adopted for simplicity and convenience during the multi-stage model training. But, we want to remark that alleviation of data imbalance is likely an important direction for future model improvement. 

\begin{figure}[ht]
 \centering 
 \includegraphics[scale = 0.3]{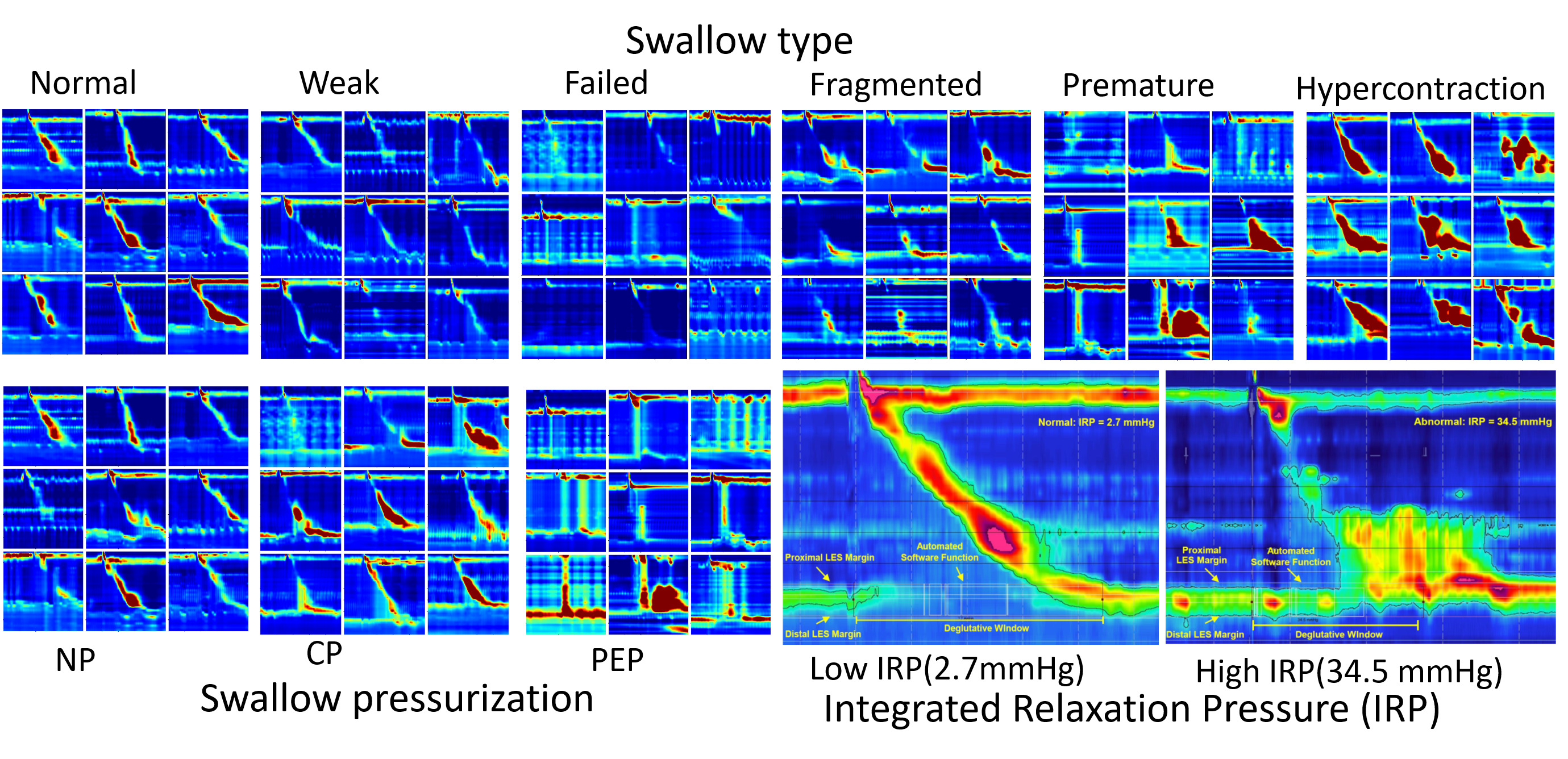}
 \caption{Explorer of HRM swallow showing the variation of pressure pattern by swallow type (Upper) and by swallow pressurization (Lower left). Lower right shows swallows with low integrated relaxation pressure (IRP) and high IRP, respectively. The calculation of IRP involves manual identifications of several landmarks. (Part of the figure is reproduced with permission from~\citep{kou2021deep}.)}
 \label{fig-dataset}
\end{figure}

\subsection{swallow-level models: 3 convolutional neural network (CNN) models}

The swallow-level models include three individual models: prediction of 3-category swallow pressurization, 6-category swallow type, and integrated relaxation pressure (IRP). These models are referred to here as \textit{swallow-type model}, \textit{swallow-pressurization model}, and \textit{irp model}, respectively. The swallow-type model and swallow-pressurization model are multi-category classification problems, whereas the IRP model is a regression problem to predict non-negative values. Inspired by the previous work and image-like patterns illustrated in Figure.~\ref{fig-dataset}, all of the three models are  based on convolutional neural networks (CNN)~\citep{kou2021deep}. Each of these three CNN models were trained individually, though they share the same training/validation/testing datasets.  Implementation of these models leveraged several libraries including \textit{tensorflow} and \textit{keras}. The training was performed on Northwestern University's Quest, a high performance computing cluster.

One adaptation required of the irp model is that IRP is typically examined with reference to cut-off values from domain knowledge, such as CC algorithm. Hence predictions near the cut-off values will be more likely to impact the ultimate diagnosis. Therefore, for the IRP model, a weighted loss function with high sensitivity near the cut-off region is introduced. Let $m$ be the size of dataset, and $X=\{X_I, I \in (1,2,...,m)\}$, $y=\{y_I, I \in (1,2,..,m)\}$ denote the features and labels, respectively. The loss function associated with one single data-point $(X_I, y_I)$ is as below
\begin{equation}
\left(\frac{1}{\lambda + 1} + \frac{\lambda}{\lambda + 1}e^{\frac{-(y_I - y_o)^2}{2y_o^2}}\right)(f(X_I) - y_I)^2 \label{eqn_irp_loss}
\end{equation}
Where, $y_o$ is the nominal cut-off value. $\lambda$ is the weight, and $f(X_I)$ is the predicted IRP from input data $X_I$. It can be shown that when $\lambda =0$ or $y_I/y_o$ is far away from 1.0, Eq.~\ref{eqn_irp_loss} becomes the standard L2-loss. Those parameters are set as $\lambda = 5.0, y_o=15.0$ in the IRP model, based on model testing.

\subsection{Study level models}
\label{sec_model}

The study-level model takes the outcomes from multiple swallow-level predictions as inputs. Hence, one issue for the development of study-level models relates to the choice of specific input features. Again, motivated by the CC approach, we adopt input features of study-level models based on \textit{statistics} of swallow-level outcomes/predictions, instead of simple stacking of per-swallow outcomes. Conversion to statistics also helps maintain symmetry with respect to swallow order and likely suppress certain errors from swallow-level predictions via statistical filtering. The specific input features will be discussed later.

Another question regarding inputs for study-level models is what source of input data to use for model training, since we could adopt \textit{ground-truth} swallow-level outcomes or model-predicted swallow-level outcomes. The choice of former will make study-level models and swallow-level models independent from each other in training, but lead to low reliability in quantifying and comparing performance of hierarchical models in validation or test. Hence in this work, we choose model-predicted swallow-level outcomes to form the input data of study-level models in both training and validation. Validity of this choice is also supported by the high accuracy of all the swallow-level models, discussed later. This choice also likely increases the reliability in model selection oriented for a two-stage hierarchical pipeline.

\subsubsection{Model family: rule-based model, Xgboost, and artificial neural network (ANN) }

\begin{figure}[ht]
	\centering 
	\includegraphics[scale = 0.2]{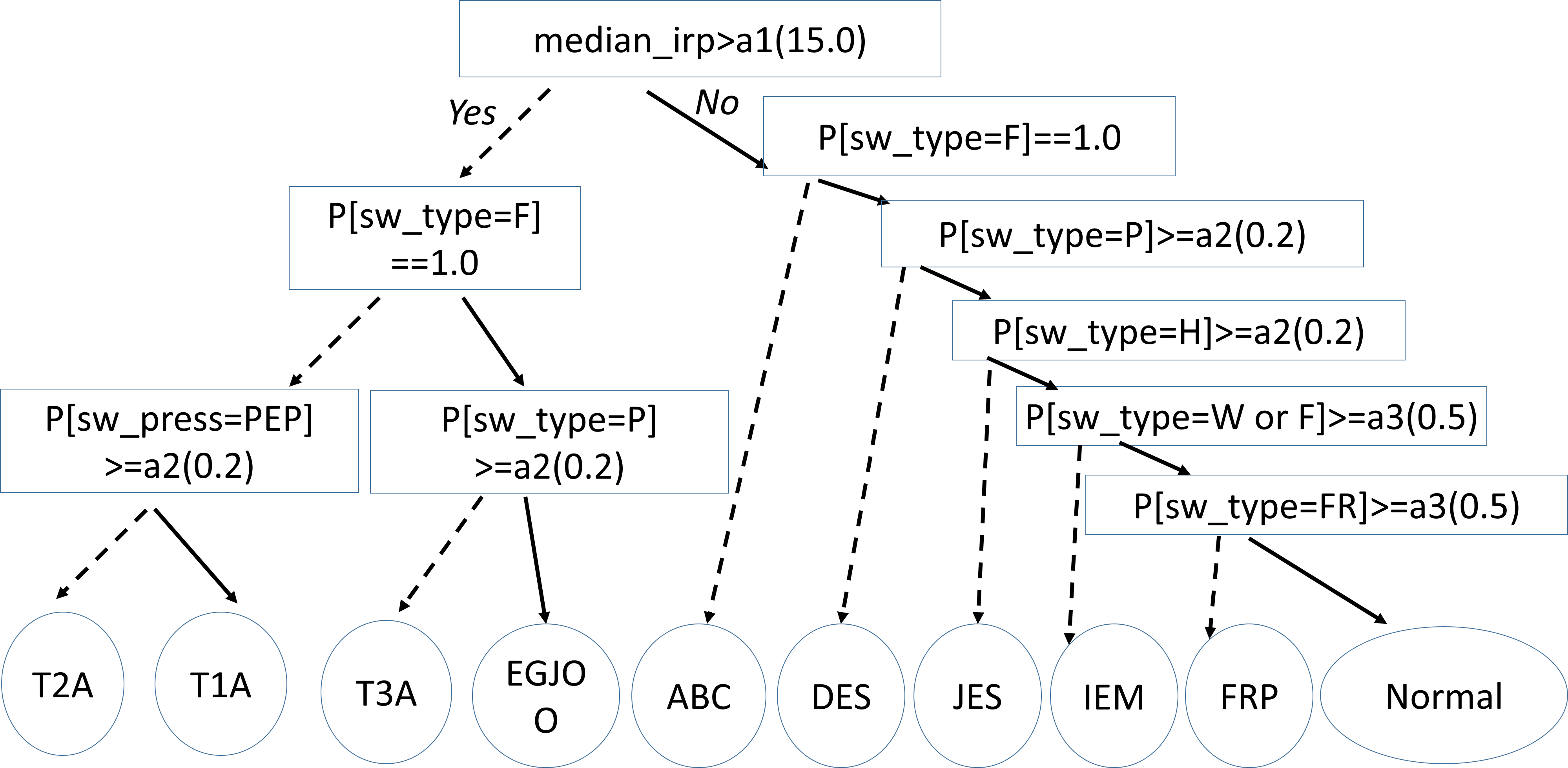}
	\caption{Illustration of the manually-designed decision rules based on Chicago Classification, referred to as the rule-based model. The $P[E]$ denotes the probability of event, $E$. The dashed-line arrow represents \textit{Yes} branch, whereas the solid-line arrow the \textit{No} branch under each decision condition. Three key parameters, $(a_1,a_2,a_3)$ were set as $(15.0, 0.2,0.5)$ in the selected model. Like the Chicago Classification, the output of this rule-based model is of 10 category, which is then merged to 8 categories (DES was merged into T3A and FRP was merged into IEM). Further details on labels could be found in Table.~\ref{table_data_1}. }
	\label{fig-rule}
\end{figure}

Instead of starting from scratch, we constructed the first family of candidate models with inspiration from current expert knowledge, i.e. the CC algorithm. Similar to the CC algorithm, we manually designed a rule based on a decision tree with outcomes from swallow-level models. Hence, this model is referred to as the \textit{rule-based model}, as illustrated in Figure.~\ref{fig-rule}. We used a grid search to develop a rule-based model and compared that model to the CC algorithm. To formalize this comparison, we established the CC algorithm by setting several key parameters at branching cuts to their values from the CC algorithm, referred to as \textit{nominal values}. These nominal values, shown in Figure.~\ref{fig-rule}, are $[a_1,a_2,a_3] = [15.0,0.2,0.5]$. $a_1$ denotes the cut-off IRP value. $a_2$ denotes both the probability of swallows with swallow type as premature and the probability of swallows with swallow pressurization as pan-esophageal pressurization. $a_3$ denotes both the probability of swallows with swallow type as weak or failed, and the probability of swallows with swallow type as fragmented. Then we conducted a grid search within intervals, $a_1 \in [12,17], a_2 \in [0.1,0.3], a_3 \in [0.4,0.6]$. This grid search found optimal values of parameters to be $[14.5,0.1,0.43]$, based on training dataset, however those optimal values only led to minimal improvement of accuracy in validation dataset relative to the nominal values from the CC algorithm (0.762 based on the nominal values vs 0.766 based on the optimal values). Therefore, for the following discussion, we selected the rule-based model with nominal values, as shown in Figure.~\ref{fig-rule}. 

Note that, the above rule-based model not only provided one candidate study-level model, but also could augment the input features with its prediction. Hence we provide two sets of input features that were used in model development and evaluations. The first set is referred to as \textit{original features}, including 13 features in 3 groups.
\begin{itemize}
    \item \textit{irp-max-normalized, irp-min-normalized, irp-median-normalized, irp-mean-normalized,}
    \item \textit{pressurization-0-prob, pressurization-1-prob, pressurization-2-prob,}
    \item \textit{type-0-prob, type-1-prob, type-2-prob, type-3-prob, type-4-prob, type-5-prob, type-6-prob.}
\end{itemize}
Where, \textit{irp-max-normalized = maximal value of irp/15.0} and the similar rule applies for other statistics of IRP-related features. The feature, \textit{pressurization-i-prob} denotes \textit{probability of the event that swallow-pressurization's id is i}, and similarly, \textit{type-i-prob} denotes \textit{probability of the event that swallow-type's id is i}.  The second set include the original features and \textit{rule-label}, the prediction from rule-based model, hence 14 features in total. This set of input features is referred to as \textit{augmented features}.

For convenience, we denote the size of dataset as $m$, vectors of input features as $\{X_I, 1\leq I \leq m \}$, and the output labels as $\{y_I \in[1,..,K], 1 \leq I \leq m\}$ with $K$ as the number of different categories. Then, at $t$-iteration, we obtain
\begin{align}
TL^t(\theta) &= \sum_{I=1}^m l(y_I,\hat{y}_I^t) + \Omega(\theta) \label{eqn_bst_1} \\
\hat{y}_I^t& = \sum_{n=0}^t(f_n(X_I)) = \hat{y}_I^{t-1} + f_t(X_I; \theta^t) \label{eqn_bst_model} \\
\hat{y}_I^0 &= b(X_I)=b_I \label{eqn_bst_bl}
\end{align}
In Eq.~\ref{eqn_bst_1}, $TL^t(\theta)$ is the total loss function at iteration $t$; $\Omega(\theta)$ denotes the loss from regularization.  $l(y_I,\hat{y}_I^t)$ is the cross-entropy loss, with $\hat{y}_I^t =\{\hat{y}_I^{t,k},1 \leq k \leq K \}$ as the soft probability of each class (i.e. the output of softmax) predicted at step/iteration $t$. Eq.~\ref{eqn_bst_model} illustrates the boosting procedure from iteration $t-1$ to iteration $t$.  In Eq.~\ref{eqn_bst_bl}, $\hat{y}_I^0$ denotes the base learner, i.e. the leaner at iteration 0. Note that the base learner could be set to be any pre-trained model, denoted as $b_I$. Instead of an explicit form, $b(X_I)$, the base model only needs to know the predicted results for each data point, hence a vector of predictions, like $b_I$ is enough. In this work, we are free to choose $b_I$ as the prediction of the rule-based model or random initialization.

The third candidate of model family is the artificial neural network (ANN) model. By incorporating hidden layers, ANNs are also capable of modeling complex relationships between input features and output~\cite{goodfellow2016deep}. Motivated by the threshold-based tree model, we adopt \textit{relu} as the activation functions of intermediate layers, whereas we adopt the \textit{softmax} as the final output layer for this multi-category (i.e. 8-category) classification problem. We adopt fully-connected architecture and vary the width and depth of networks during model selection.

\subsubsection{Blended model: Model balancing/averaging}

From model training and selection, we are likely to obtain multiple well-performing models, one from each family, referred to here as sub-models. A natural question is how to balance or combine sub-models to yield a single blended model that performs better than any individual sub-model. A general strategy of model balancing suitable for different model families is still the focus of extensive theoretical research~\cite{oza2008classifier,hoeting1999bayesian,wasserman2000bayesian}. Bayesian model averaging offers one unified theoretical framework, but it typically require the sub-models from the same family to facilitate the factorization of prior probability~\cite{hoeting1999bayesian,wasserman2000bayesian}. For sub-models from multiple different families, a simple weighted average, such as equal weights or biased weights per-model, could be adopted, though this approach lacks the formal probabilistic justification as Bayesian averaging~\cite{oza2008classifier}. Thus, to account for the limitations for both approaches, we adopt a model-agnostic framework of model balancing with motivation from Bayesian principles.

Following previously introduced notations, we denote the $I$-th data-point as $(X_I, y_I)$, where $y_I$ is the output as a single (i.e. the most likely) index among $K$-category labels. This kind of output is referred to as \textit{single-index} output. In contrary, another type of model output could be based on \textit{probability} of each category, denoted as $y^k_I, 1\leq k \leq K$. This second type of output is referred to as \textit{soft-probability} output, and contains more information. Note that, conversion from the single-index output to soft-probability output could be done via one-hot encoding. Hence, we assume we have $J$ pre-trained models, which all generate soft-probability output. Specifically, for a given input $X$, we denote the output of $j$th pre-trained model as, $M_j^k = M_j(y=k|X), 1\leq k \leq K$. Then the blended model, denoted as a functional form $B(y=k|X), 1\leq k \leq K$, could be approximated as below.

\begin{align}
B(y=k|X) &\simeq \sum^J_{j=1}B(y=k, M_j(y=k)|X) \label{eqn_bm_1} \\
         &\simeq \sum^J_{j=1}B(y=k|M_j(y=k))(M_j(y=k|X)) \label{eqn_bm_2} \\
         &\simeq C\sum^J_{j=1}Prob(y=k|M_j(y=k))(M_j(y=k|X)). \label{eqn_bm_3} 
\end{align}
Eq.~\eqref{eqn_bm_1} assumes a form of marginalization over $j$ of a joint probability distribution $B(y=k, E_j |X)$, where $E_j=M_j(y=k|X)$. Eq.~\eqref{eqn_bm_2} is a model-agnostic assumption so that the blended model only directly depends on outputs of pre-trained models, not their specific forms or input $X$. Eq.~\eqref{eqn_bm_3} could be considered as an approximation of $B(y=k|M_j(y=k))$ by $Prob(y=k|M_j(y=k))$, which can be identified as the precision score of the $j$-th model in predicting the $k$-th category. The constant $C$ is a normalization factor such that $\sum^K_{k=1}B(y=k|X)=1$. Hence, the proposed framework of model balancing could be summarized as below.
\begin{align}
B(y=k|X) &\simeq C\sum^J_{j=1}PS_j^kM_j(y=k|X), \label{eqn_bm_4} 
\end{align}
where $PS_j^k$, the precision score of the $j$-th model in predicting the $k$-th category, acts as the weight of individual model's soft-probability output, $M_j(y=k|X)$. We remark Eq.~\eqref{eqn_bm_4} is a model-agnostic heuristic rule for model balancing, motivated by Bayesian principles. Unlike the simple weighted averaging per model~\cite{oza2008classifier}, the weight of each model here also varies across categories based on the precision score. Intuitively, the precision score represents the \textit{credit score} of each model in $correctly$ predicting each category and hence can naturally be adopted as the weight.

\subsubsection{Other details on model comparison and selection}
\label{sec_others} 

For study-level models, we first train and compare models within each family. Model training and selection were based on performance on training and validation datasets. In particular, the overall accuracy on the validation dataset was the criterion to select the optimal candidate model in each family. The held-out test dataset was then used to evaluate the selected model for details. After selecting candidate model from each family, we then conduct model balancing to obtain blended models via different combinations. Similarly, the blended models were then compared based on their performance on validation dataset and evaluated using test dataset.

Regarding the implementation, study-level models were developed based on several python libraries including \textit{tensorflow}~\citep{abadi2016tensorflow} and \textit{keras}~\citep{chollet2015keras} for ANN models and \textit{Xgboost} library for xgboost models~\cite{chen2016xgboost}.

\section{Results and Discussions}
\label{sec_results} 
\subsection{Swallow-level models}
\begin{figure}[htb]
	\centering 
	\includegraphics[scale = 0.25]{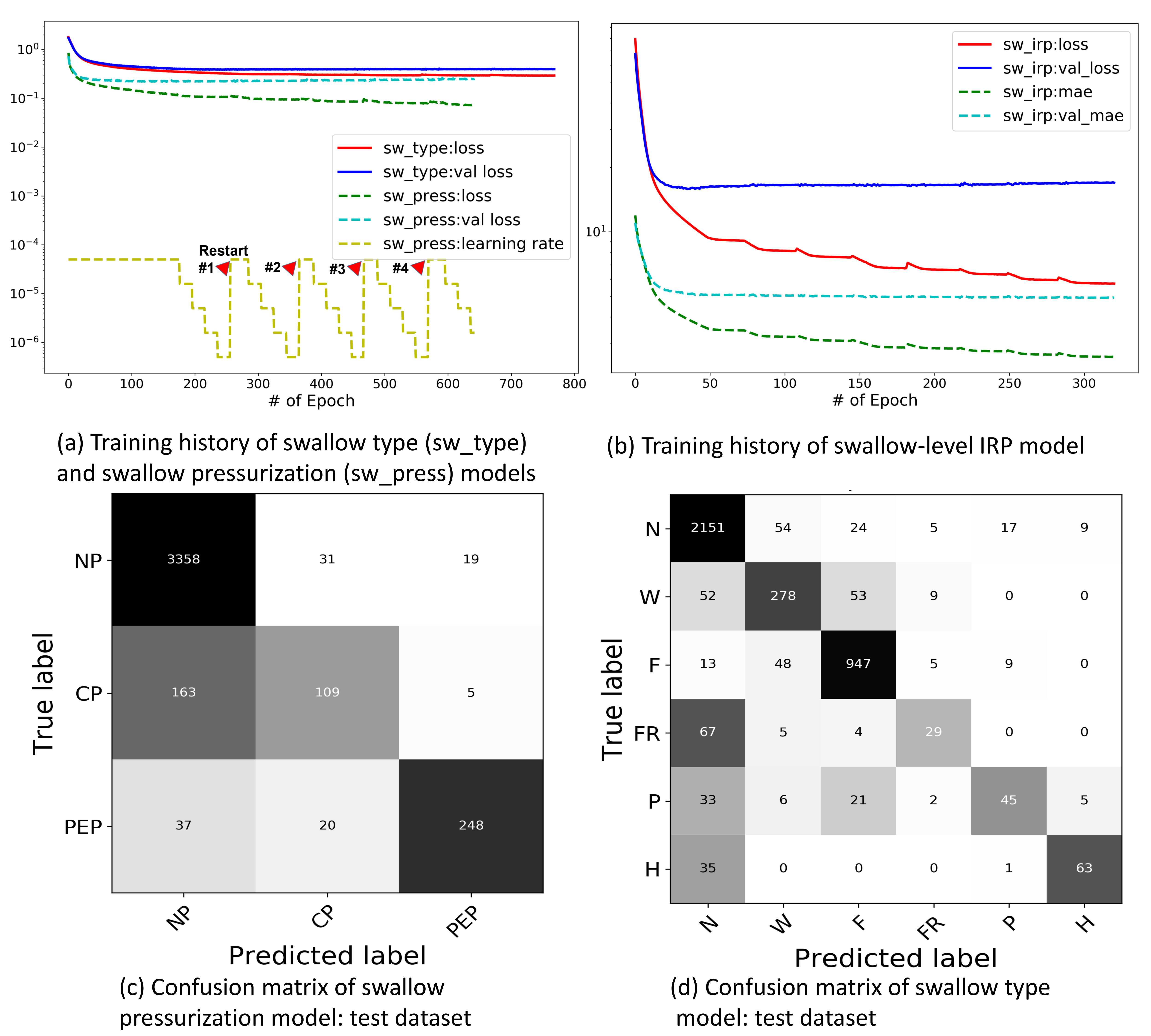}
	\caption{Training history of the swallow-level AI models and their evaluation based on Confusion matrix on testing dataset. (a)}
	\label{fig-swallow-model}
\end{figure}

Based on extensive experimentation that varied the number of network layers and network width, a similar architecture of convolutional neural networks (CNN) for both the swallow-type model and swallow-pressurization model was obtained, as listed in Table.~\ref{table_sw_type}.
For the regression-type IRP model, more convolutional layers were found to be necessary, which is likely related to the challenge of landmark identification to extract the region of interest, as illustrated in Figure.~\ref{fig-dataset} (the lower left panel). The CNN architecture of the IRP model is listed in Table.~\ref{table_irp}. The training history of the three swallow-level models were included in Figure.~\ref{fig-swallow-model} (a) and (b). The training process involved a first fresh run, followed by multiple restart runs based on saved model weights of the best validation accuracy. As Figure.~\ref{fig-swallow-model} (a) and (b) shows, although the training loss continued to decrease with the number of epochs, the validation loss achieved a plateau during the restart runs, indicating a likely convergence. 

The final trained models were then evaluated on training/validation/test datasets, with results listed in Table.~\ref{table_sw_acc}. The accuracy of swallow-type model and swallow-pressurizaton models were 0.88 (test) and 0.93 (test), respectively, which, to our knowledge, represents the best performance among AI models on these tasks. The mean absolute error (MAE) of the IRP model is 4.5 (mmHg), which, to our knowledge, represents the first AI model to automatically predict IRP values from raw pressure measurements. However, the IRP model, compared with the swallow-type and swallow-pressurization models, showed a large gap between training loss and validation/test loss. This might indicate an apparent over-fitting and further model improvement might be needed in future. Detailed per-class evaluation of swallow-type and swallow-pressurization models are conducted based on the test dataset, as shown in Table.~\ref{table_sw_performance} and Figure~\ref{fig-swallow-model}. For the swallow-type model, predictions appear to be most accurate among Normal (N) and Failed (F) groups, whereas predictions among Premature (P) and Fragmented (FR) groups tend to have the lowest accuracy. One factor is related to sample imbalance, as Normal and Failed groups comprise the majority of observations in the data. Another factor relates to the degree of similarity of patterns among certain swallow types, such as Normal, Fragmented and Premature, as illustrated by confusion matrix in Figure~\ref{fig-swallow-model} (d). Similarly, for the swallow-pressurization model, the differences in precision/recall/F1-score across classes is also likely related to sample imbalance.

\begin{table}[h]
 \caption{Layers in convolution neural network (CNN) model on predicting swallow pressurization and swallow type. The first dimension of \textit{Output shape}, labeled as \textit{None}, corresponds to the size of mini-batch. The last dimension denotes the number of channels, in which {C=2, K=3} for the CNN model of swallow pressurization and {C=1, K=6} for the CNN model of swallow type. Details of layer type could be found from Keras documentation~\citep{chollet2015keras}}
 \centering
\begin{tabular}{l | l  }
 \hline \hline
 Layer type & Output Shape \\ [1ex]
 \hline
 InputLayer  	  & (None, 36, 240, 1) 	\\
 Conv2D+BN           & (None, 35, 79, 8C)     \\
 MaxPooling       & (None, 34, 38, 8C)  \\
 Conv2D+BN           & (None, 33, 36, 16C)    \\
 MaxPooling       &  (None, 32, 34, 16C)   \\  
 Conv2D+BN  	      &  (None, 30, 32, 32C)    \\
 MaxPooling 	  & (None, 14, 15, 32C)  \\
 Conv2D+BN           &  (None, 12, 13, 64C)  \\
 MaxPooling       & (None, 5, 6, 64C)  \\
 global average pooling2d & (None, 64C) \\
 Dense			&	(None, 64C) \\
 Dense			&	(None, 64C) \\
 Dense		    &     (None, K) \\
 \hline
 \end{tabular}
 \label{table_sw_type}
\end{table}

\begin{table}[h]
	\caption{Layers in convolution neural network(CNN) model on predicting integrated relaxation pressure (IRP). The first dimension of \textit{Output shape}, labeled as \textit{None}, corresponds to the size of mini-batch. The last dimension denotes the number of channels. batch normalization(BN) is adopted. Details of layer type could be found from Keras documentation~\citep{chollet2015keras}}
	\centering
	\begin{tabular}{l | l  }
		\hline \hline
		Layer type & Output Shape \\ [1ex]
		\hline
		InputLayer  	  & (None, 36, 240, 1) 	\\
		Conv2D+BN           & (None, 35, 79, 16)     \\
		MaxPooling       & (None, 34, 38, 16)  \\
		Conv2D+BN           & (None, 33, 36, 32)    \\
		MaxPooling       &  (None, 32, 34, 32)   \\  
		Conv2D+BN  	      &  (None, 30, 32, 64)    \\
		MaxPooling 	  & (None, 14, 15, 64)  \\
		Conv2D+BN           &  (None, 12, 13, 64)  \\
		MaxPooling       & (None, 5, 6, 64)  \\
		Conv2D+BN           &  (None, 3, 4, 128)  \\
		MaxPooling       & (None, 3, 4, 128)  \\
		global average pooling2d & (None, 128) \\
		Dense			&	(None, 128) \\
		Dense			&	(None, 128) \\
		Dense		    &     (None, 1) \\
		\hline
	\end{tabular}
	\label{table_irp}
\end{table}

\begin{table}[htb]
 \caption{Training and evaluation results for the three swallow-level models: the swallow-type model, the swallow-pressurization model, and the IRP model based on various metrics. \textit{c-ent} denotes categorical cross-entropy loss; $acc$ denotes accuracy on multi-class predictions; \textit{m-mse} denotes the modified mean-square-error loss from Eq.~\ref{eqn_irp_loss}; $mae$ denotes mean absolute error.}
 \centering
\begin{tabular}{l | l | l | l | l }
 \hline \hline
 CNN model & metric& training  & validation  & testing \\ [1ex]
 \hline
  swallow type &c-ent& 0.29 &0.38&0.35\\
  swallow type &acc& 0.90 &0.88&0.88\\
  swallow pressurization & c-ent& 0.1 &0.29&0.23\\
  swallow pressurization & acc & 0.96 &0.92 &0.93\\
  IRP & m-mse& 30.98 &60.24&68.74\\
  IRP & mae & 2.57 &4.71 &4.49\\
 \hline
 \end{tabular}
 \label{table_sw_acc}
\end{table}

\begin{table}[!htb]
	\caption{Detailed performance in swallow level models in testing dataset.}
	\centering
	\begin{tabular}{c | c | c| c |c | c }
		\hline \hline
		Category & label  & precision & recall &F1& Support\\ [1ex]		      
		\hline
		swallow type	&N  & 0.91 &    0.95 &    0.93&     2260 	\\
		&W  & 0.71 &    0.71&      0.71 &      392	\\
		&F  & 0.90 &    0.93 &    0.91 &     1022 		\\
		&FR  &0.58&      0.28 &     0.37&      105	\\
		&P  & 0.62 &    0.40 &     0.49&      112		\\
		&H  & 0.82 &    0.64 &     0.72 &      99		\\
		\hline
		swallow 		&NP   & 0.94 &     0.99&    0.96 &     3408		\\
		pressurization &CP   & 0.68&    0.39 &      0.50 &      277		\\
		&PEP   &0.91  &    0.81 &     0.86&      305	\\
		\hline
	\end{tabular}
	\label{table_sw_performance}
\end{table}

\subsection{Study-level models}

\subsubsection{Boosting tree model: xgboost}
The xgboost model was built to predict the soft-probability output of the 8-category study label, following a one-vs-other approach. The number of iterations/estimators, $n$, was determined by monitoring the change of validation loss, with the learning rate set as $0.3$. Hence, each trained model was actually composed of $8*n$ decision trees. For model comparison and selection, the variations considered includes three factors: input features (original features or augmented features), depth of trees, as well as consideration of rule-based model as the base learner. Detailed model comparison is discussed in the appendix. In particular, overall accuracy of various models are listed in Table.~\ref{table_xgboost_acc}, which indicates a similar performance was found across all the models. Based on validation accuracy, the final selected model was \textit{xgb-5}, with depth of 4, with rule-based model as the base learner, and with augmented features as inputs. This may suggest that model augmentation via a base learner as well as input augmentation could be a way to improve the performance. The overall accuracy of the model \textit{xgb-5} across training/validation/testing datasets are 0.94/0.79/0.81. The detailed per-class evaluation of the model on the test dataset is listed in Table.~\ref{table_xgboost_score}. Although NEM, the most common label, is predicted with the highest accuracy, EGJOO, which is the second most common label, has a lower accuracy than some minority groups such as IEM and T2A. This is because EGJOO represents one of the most inconclusive diagnoses in Chicago Classification, share a high similarity/overlap with many other groups. The observation is also consistent with clinical impression on EGJOO cases, where data from other procedures are often needed for confirmation.

\begin{table}[!htb]
	\caption{Detailed per-class scores of the model \textit{xgb-5} on testing dataset. Model \textit{xgb-5} is the selected xgboost model from Table.~\ref{table_xgboost_score}. Its overall accuracy across training/validation/test datasets are 0.94/0.79/0.81.}
	\centering
	\begin{tabular}{ c | c| c |c | c }
		\hline \hline
		Label  & precision & recall &F1& Support\\ [1ex]		      
		\hline
		ABC  &0.64      &0.69      &0.67       & 13	\\
		T1A &  0.50      &0.60      &0.55       & 10	\\
		T2A  & 0.88      &0.83      &0.86      &  18		\\
		T3A  &  0.83      &0.33     & 0.48       & 15	\\
		EGJOO  & 0.62      &0.73    &  0.67      &  48	\\
		JES  &1.00      &0.33      &0.50        & 6	\\
		NEM  &  0.89      &0.89     & 0.89      & 119	\\
		IEM  & 0.78      &0.84      &0.81       & 37	\\		
		\hline
	\end{tabular}
	\label{table_xgboost_score}
\end{table}


\subsubsection{ANN model}
To train and select our ANN model, we varied the depth and width of hidden layers. Cases with different input features, i.e. original features or augmented features, were also compared. The generic form of model architecture is listed in Table.~\ref{table_ann_model}, where the hidden layers were all fully-connected (dense) layers with $relu$ activation. Detailed discussions and results on model comparison can be seen in the appendix, in particular, Table.~\ref{table_xgboost_acc}. The selected ANN model based on Table.~\ref{table_ann_acc} is \textit{ANN-1}, with 6 dense layers and augmented features as input. The model's overall accuracy across training/validation/test datasets are 0.84/0.77/0.80, less over-fitting than xgboost model, \textit{xgb-5}. The per-class precision/recall/F1 scores are listed in Table.~\ref{table_ann_score}, which is very similar to results of xgboost model, \textit{xgb-5}.

\begin{table}[h]
	\caption{Architecture of one example of the ANN model with 6 dense layers. As the input dimension (Nx=14 for augmented features and Nx=13 for original features) and output dimension (8) is fixed by the problem, width of neural network is controlled by factor K for variation in each hidden layer. All the dense layers adopt $relu$ activation except the output dense layer, which adopts $softmax$ activation.}
	\centering
	\begin{tabular}{l | l  }
		\hline \hline
		Layer type & Output Shape \\ [1ex]
		\hline
		Flatten  	&	(None, Nx) \\
		Dense x 2			&	(None, 50K) \\
		Dense x 3			&	(None, 25K) \\
		Dense		    &     (None, 8) \\
		\hline
	\end{tabular}
	\label{table_ann_model}
\end{table}

\begin{table}[!htb]
	\caption{Detailed scores of model ANN-1 (see Table.~\ref{table_ann_acc}) on test dataset. The overall accuracy across training/validation/test datasets are 0.84/0.77/0.80.}
	\centering
	\begin{tabular}{ c | c| c |c | c }
		\hline \hline
		Label  & precision & recall &F1& Support\\ [1ex]		      
		\hline
		ABC  &0.64      &0.69      &0.67       & 13	\\
		T1A &  0.50      &0.60      &0.55       & 10	\\
		T2A  & 0.89      &0.89      &0.89     &  18		\\
		T3A  &  0.80     &  0.53     &  0.64        & 15	\\
		EGJOO  & 0.70    & 0.69     & 0.69      &  48	\\
		JES  &1.00      &0.33      &0.50        & 6	\\
		NEM  &  0.88      &0.92     & 0.90      & 119	\\
		IEM  & 0.76      & 0.78     & 0.77        & 37	\\		
		\hline
	\end{tabular}
	\label{table_ann_score}
\end{table}


\subsubsection{Blended model: balancing the selected rule-based model, xgboost model, and ANN model}
The model fitting and selection process outlined above produced three pre-trained models from each family: the rule-based model, the xgboost model xgb-5, and the ANN model ANN-1. Notice that for the rule-based model, the output is single index of the study label, whereas for xgb-5 and ANN-1, the model outputs include two types, 8-class soft-probability output and single-index output derived from soft-probability outputs. Based on Eq.~\eqref{eqn_bm_4}, model balancing based on these two different outputs were conducted, with the weights as precision scores from the training dataset. Different combinations of the selected models were considered. The accuracy score from Top-2 predictions was defined as probability that the true label is contained in the first two most likely labels predicted from the model.Overall accuracy score, including that of the top-2 predictions from validation dataset were compared, as listed in Table.~\ref{table_mix_acc}. Although the blended model using the single-index output shows a slight higher validation accuracy from top-1 prediction, its score from top-2 predictions are much lower. Hence, with respect to accuracy of the top-2 predictions, model \textit{xgb+ann-1} was selected from the blended models.

\begin{figure}[h!]
	\centering 
	\includegraphics[scale = 0.28]{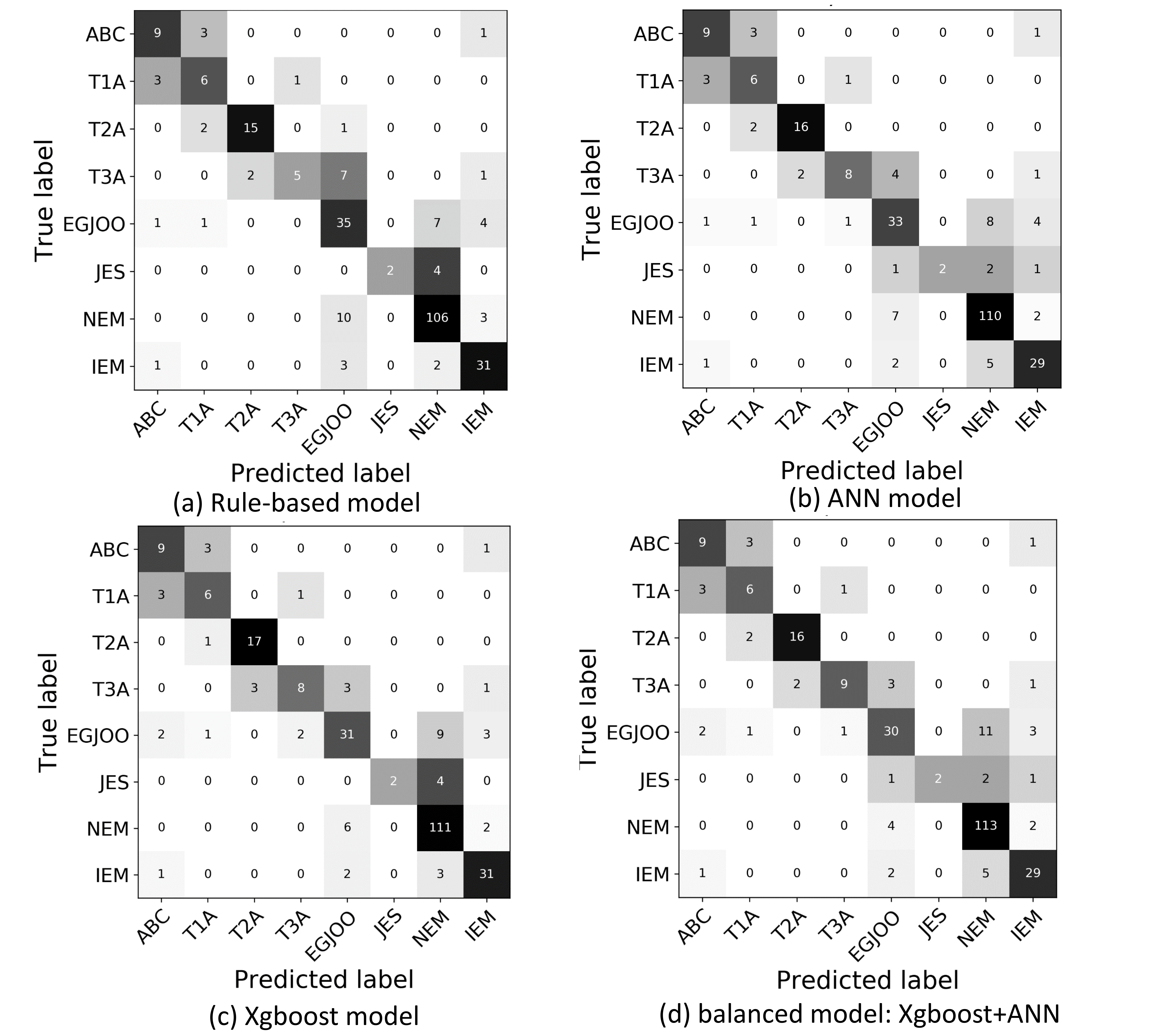}
	\caption{Evaluation of study-level models selected from each family based on confusion matrix on the testing dataset}
	\label{fig-study-model}
\end{figure}

In addition, we evaluated and compared the overall accuracy of sub-models across all the three datasets, including the top-2 predictions from the test dataset. The best model according to that criterion was the xgboost model \textit{xgb-5}, though ANN model \textit{ann-1} and blended model \textit{bst+ann-1} achieved very similar scores. A further comparison can be seen from the confusion matrix plot on the test dataset, as shown in Figure.~\ref{fig-study-model}. All of the models share very similar confusion patterns, probably related to augmentation of rule-based model during model development. This may also explain why the blended model did not improve the performance, since all the pre-trained sub-models share almost identical strength and weakness. However, compared with the rule-based model, all the other three models yielded improved accuracy. More importantly, they present a higher accuracy from the top-2 predictions (above 0.91).

\begin{table}[h!]
	\caption{Overall accuracy of all the selected models, one from each family, for comparison. \textit{rule} is the rule-based model in Figure.~\ref{fig-rule}. The \textit{test-top1} is the score based on top-1 prediction of test dataset, whereas the \textit{test-top2} is the score based on top-2 predictions. }
	\centering
	\begin{tabular}{l | l | l | l | l  }
		\hline \hline
		Model & training  & validation &test-top1 & test-top2 \\ [1ex]
		\hline
		rule & 0.82 &0.76 & 0.79 & NA \\		
		xgb-5 & 0.94&0.79 & 0.81 & 0.92 \\
		ann-1 & 0.84 & 0.77 & 0.80 & 0.92 \\
		
		\hline
		xgb+ann-1 & 0.86&0.77 & 0.80 & 0.91	\\	
		\hline
	\end{tabular}
	\label{table_all_acc}
\end{table}

%

\begin{figure}[h!]
	\centering 
	\includegraphics[scale = 0.5]{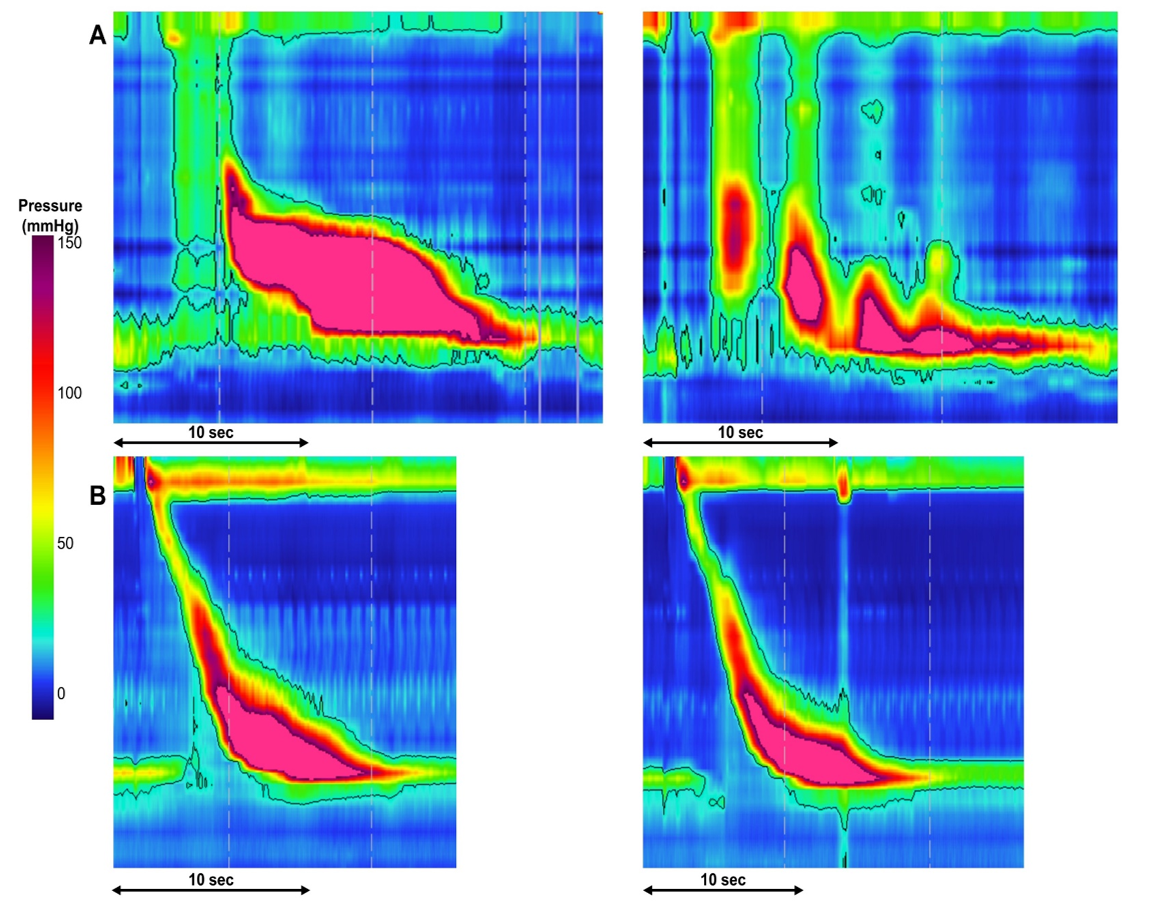}
	\caption{Examples of misclassified studies. Two test swallows from two patients (A and B) are included. Patient A had a true label of EGJ outflow obstruction (EGJOO), while was predicted to have type III (spastic) achalasia (T3A). Patient B had a true label of EGJOO, while was predicted to have normal motility (NEM). }
	\label{fig-mis-classification}
\end{figure}
\subsubsection{Misclassified cases: EGJOO}

As demonstrated in Figure.~\ref{fig-study-model}, the greatest proportion of model misclassifications is related to the diagnosis of EGJOO. Two EGJOO patients (referred to as A and B), as shown in Figure.~\ref{fig-mis-classification}, are discussed as examples of model mis-classification. Patient A was predicted to have type III (spastic) achalasia (T3A). Its EGJOO diagnosis was based on 9/10 swallows having normal distal latencies (i.e. not spastic swallows), such as the swallow on the left with a distal latency of 6.5 seconds.  However 1/10 swallows were spasm (premature) with distal latency values less than 4.5 seconds, such as the swallow on the right.  The updated Chicago Classification version 4.0 (CCv4.0) would label this study as EGJOO with spastic features. Patient B was predicted to have normal motility. The supine swallows had a median integrated relaxation pressure (IRP) values of 20 mmHg (such as the supine swallow on the left), though the IRP normalized in the upright position, as in the swallow on the right with an IRP of 10 mmHg.  Thus the predicted value (NEM) was actually more correct, as application of the updated CCv4.0 criteria would label this study as normal. In summary, certain misclassified studies could be borderline cases, where firm lines between diagnoses/classifications may not truly reflect specific disease states, especially among the EGJOO classification that has been recognized as a non-specific HRM pattern (i.e. it can represent various clinical diagnoses). In fact, the recent update of the Chicago Classification (version 4.0) recommends that all manometric EGJOO classifications be considered clinically 'inconclusive' and other diagnostic tests should be pursued to clarify the diagnosis prior to initiating treatment of a potential motility disorder. Overall, future work remains needed to further examine the clinical characteristics and clinical significance of these misclassified cases, as AI models such as these may improve identification of specific esophageal motility phenotypes.

\section{Conclusion}

This article proposed a multi-stage hierarchical modeling framework in order to identify both swallow-level outcomes and study-level diagnoses. This framework included generating multiple sub-models for diagnosis and balancing their output. We found that sub-models and their blended output predicted diagnosis with a very good accuracy. The modeling framework, though complex in architecture, could be re-packaged as a single predictive tool on raw manometry data. In particular, the modeling framework could automatically identify 6-category swallow-type and 3-category swallow-pressurization labels with accuracy of 0.88 (test) and 0.93(test), and an IRP value with mean absolute error of 4.49(mmHg) (test), showing the best performance on related tasks. Moreover, the modeling framework was also able to predict 8-category study-level diagnosis with top-1 accuracy of 0.81 (test) and top-2 accuracy of 0.92 (test), which , to our knowledge, is the first-of-the-kind model that targets Chicago classification diagnosis from raw manometry data.

There are two features in the proposed modeling framework that are generalizable for similar applications. First, its hierarchical pipeline enables the prediction of both study diagnosis and variables relevant to that diagnosis: swallow-level outcomes. The two-level architecture not only provides more information during the predictive stage, but also alleviates the issue of small sample size and large data dimensionality to support a direct study-level model during the training stage. Second, the framework incorporates domain knowledge during model development. The presented approach leveraged a  rule-based model from the Chicago Classification scheme. This rule-based model was manually designed and then used to augment input features, serve as the base learner for boosted models, and join with other models in model balancing. Evidenced from model comparison and selection, feature and model augmentation via the base learner did improve the performance of the xgboost model, yielding model \textit{xgb-5} as the optimal model (see Table.~\ref{table_xgboost_acc}). Model augmentation via model balancing, however, did not improve the performance in the current work, probably due to a high similarity in predictive strength among all the sub-models. But the methodology of model balancing could be extended to more generic tasks, such as multi-stage multi-modal modeling framework. For example, in building a diagnosis platform with both HRM data and FLIP data, a separate model on each data mode (i.e. FLIP data mode and HRM data mode) could be developed \textit{separately and independently}. Then blended models with the same model-agnostic methodology of model balancing could be developed, which could combine the strength of each data mode in detecting defining patterns among phenotypes.

Although the current work could serve as a benchmark model on predicting HRM diagnosis from raw data, two limitations exists that could be addressed in future model improvement. The first limitation is unbalanced per-class performance scores across several groups, particularly on the study-level predictions. As discussed previously, the first factor arises due to unbalanced datasets in training, and improving balance across datasets in size and distribution of labels, seems to be necessary for iterative re-training in future. The other factor may relate to the ambiguity among certain groups in the 10-category CC scheme. In particular, the EGJOO group is associated with highly variable measures that overlap with those exhibiting normal outflow, achalasia and other diagnoses based on clinical experience; this is also evidenced in Figure.~\ref{fig-study-model}. This implies interpretation-focused Chicago Classification scheme may need to be revised based on more objective underlying pathology or symptoms. A data-driven approach to guide HRM classification is of a great value to pursue in future as this could refine the classification of EGJOO into clinical phenotypes along the spectrum of dysmotility. The second limitation is reliance on three swallow-level outcomes in predicting study label. Although swallow type, swallow pressurization and IRP are three key defining outcomes evidenced by the performance in the simple rule model, including more outcomes such as the encoded pattern extracted from unsupervised auto-encoder~\cite{kou2021deep}, could likely retain more \textit{information} to improve the performance. This type of feature augmentation could be pursued in future.

{\color{black} Another limitation of the current work is the lack of mechanistic details from the model prediction. Although the two-stage model retains swallow-level predictions to characterize peristaltic pattern and contraction vigor encoded by swallow type and swallow pressurization, mechanistic details reflecting underlying physiology is absent. Incorporation of bio-physical models could offer more quantitative features. For example, Frigo et al.~\citep{frigo2017procedure}introduced continuum representation of esophageal manometry by modeling the peristaltic wave in a parametric form, which was then incorporated into an automatical procedure to identify esophageal diseases. Integrating mechanistic principles and evidence-based ML models is an valuable direction to pursue to advance both clinical applications and understanding of the physiology.} 

In conclusion, an AI-style diagnostic tool on raw HRM data to automatically derive swallow-level outcomes and study diagnosis was developed and evaluated. The proposed multi-stage hierarchical modeling framework could be extended to similar applications including multi-modal tasks.

\section*{Acknowledgement} 
This work was supported by P01 DK117824 (JEP) from the Public Health service. In addition, this work was made possible by a grant from the Northwestern Digestive Health Foundation and gifts from Joe and Nives Rizza and The Todd and Renee Schilling Charitable Fund.

\section*{Appendix: model selection from xgboost family, ANN family, and blended models}
\label{apd_selection}
Here we illustrate the detailed process of model comparison and model selection from xgboost family, ANN family and blended models. For each model family, cases with different setup or hyper-parameters were conducted and compared based on overall accuracy on validation dataset. This criteria, though simple, could be biased due to sample imbalance and/or small dataset for validation, but we remark this framework could provide a consistent guidelines on model evaluation.

For the xgboost model, the variations considered here includes three factors: input features (original 13 features vs. augmented 14 features), depth of the tree, as well as consideration of rule-based model as the base learner. Results of different cases are listed in Table.~\ref{table_xgboost_acc}. Hence, the selected optimal model is xgb-5, the model of max-depth as 5 with the rule-base model as the base learner, and the augmented feature set (i.e. 14 features) as input features.

\begin{table}[h!]
	\caption{Overall accuracy of various xgboost models for comparison and selection. The \textit{rule} denotes the rule-based model (see Figure.~\ref{fig-rule}). The model selection is based on overall accuracy on validation dataset. Hence, the selected optimal model is xgb-5, the model of max-depth as 5, with the rule-base model as the base learner, and the augmented features (i.e. 14 features) as input.}
	\centering
	\begin{tabular}{l | l | l | l | l | l  }
		\hline \hline
		Model & max depth & base learner&feature NO.& training& validation\\ [1ex]
		\hline
		xgb-1 & 3 & no&14 &0.90 & 0.78 \\
		xgb-2 & 4 & no&14 &0.94 & 0.77 \\
		xgb-3 & 5 & no&14 &0.96 & 0.78 \\
		\hline
		xgb-4 & 3 & rule&14 &0.90 & 0.78 \\
		\textbf{xgb-5}& \textbf{4} &\textbf{rule} & \textbf{14} & \textbf{0.94}& \textbf{0.79} \\
		xgb-6 & 5 & rule&14 &0.96 & 0.78 \\		
		\hline
		xgb-7 & 3 & no &13 & 0.90 & 0.78 \\
		xgb-8 & 4 & no &13 &0.90 &0.76 \\
		xgb-9 & 5 & no &13 &0.96 &0.77 \\
		\hline
		xgb-10 & 3 & rule &13 & 0.90 & 0.78\\
		xgb-11 & 4 & rule &13 & 0.94 & 0.78\\
		xgb-12 & 5 & rule &13 & 0.96 & 0.77\\		
		
	\end{tabular}
	\label{table_xgboost_acc}
\end{table}

For the ANN model, the variations of model included three factors: input features (original 13 features vs. augmented 14 features), width and depth of neural networks, as well as consideration of sample weights for class balance. Results of different cases are listed in Table.~\ref{table_ann_acc}. The selected optimal model in the ANN family was adopted as ANN-1, with depth of 6 and augmented features (i.e. 14 features) as input features.
\begin{table}[h!]
	\caption{Overall accuracy of various ANN models for comparison and selection. The width factor is the factor $K$ in Table.~\ref{table_ann_model}. Note that the model selection is based on overall accuracy on validation dataset. Hence, the optimal model in the ANN family was adopted as ANN-1, with depth of 6 and augmented features (i.e. 14 features) as input features. ANN-5 and ANN-10 utilized class weights in consideration of sample imbalance. }
	\centering
	\begin{tabular}{l | l | l | l | l | l  }
		\hline \hline
		Model &layer NO.& width factor& feature NO.& training& validation\\ [1ex]
		\hline
		\textbf{ANN-1} & \textbf{6} &\textbf{1}  &\textbf{14} &\textbf{0.84}  & \textbf{0.77}  \\
		ANN-2 & 6 & 2 &14 &0.87 & 0.75  \\
		ANN-3 & 5 & 1 &14 &0.85 & 0.77 \\
		ANN-4 & 7 & 2 &14 &0.83 & 0.74 \\
		ANN-5*& 6 & 1 &14 &0.81 & 0.72 \\
		\hline
		ANN-6 & 6 & 1 &13 & 0.85 & 0.75 \\
		ANN-7 & 6 & 2 &13 & 0.85 & 0.75 \\
		ANN-8 & 5 & 1 &13 & 0.85 & 0.74\\
		ANN-9 & 7 & 2 &13 & 0.85 & 0.75\\
		ANN-10* & 6 & 1 &13 & 0.80 & 0.71\\
		\hline
	\end{tabular}
	\label{table_ann_acc}
\end{table}
For the blended model, the variation includes different type of outputs from pre-selected models as well as various combinations of them. Specifically, the output type includes single-index output and soft-probability output during model balancing. The weights are adopted as precision scores from the training dataset. Note that for models with significant over-fitting, precision scores from the validation dataset could be used, instead.
\begin{table}[h!]
	\caption{Overall accuracy of various models for comparison and selection from model balancing. The \textit{rule} is the rule-based model illustrated. $soft prob.$ denotes the output type as 8-class soft-probability output. Both top-1 prediction and top-2 predictions of the validation dataset, labeled as \textit{valid.-top1} and \textit{valid.top-2}, respectively, was used for selection among blended models. Based on the validation-top2 score, the $xgb+ann-1$ was selected among blended models.}
	\centering
	\begin{tabular}{l | l | l | l | l  }
		\hline \hline
		Model & output type & training  & valid.-top1 & valid.-top2 \\ [1ex]
		\hline		
		xgb-5 & soft prob. & 0.94&0.79 & 0.91 \\
		ann-1 & soft prob.  &0.84 & 0.77 & 0.92 \\
		rule & single index & 0.82 &0.76 & NA \\
		
		\hline
		xgb+ann-1 & soft prob. & 0.86&0.77 & 0.92 \\
		xgb+rule-1 & soft prob. & 0.82&0.76 & 0.87 \\
		ann+rule-1 & soft prob.  &0.83 & 0.76 & 0.92 \\
		xgb+ann+rule-1 & soft prob.  & 0.83 &0.77 & 0.92 \\
		\hline
		xgb+ann-2 & single index & 0.93&0.78 & 0.84 \\
		xgb+rule-2 & single index & 0.93&0.79 & 0.85 \\
		ann+rule-2& single index &0.84 & 0.78 & 0.82 \\
		xgb+ann+rule-2 & single index & 0.87 &0.78 & 0.85 \\		
		\hline
	\end{tabular}
	\label{table_mix_acc}
\end{table}



\bibliographystyle{elsarticle-num}

\bibliography{hrm_cnn}





\end{document}